\documentclass[12pt]{spieman}  
\usepackage{amsmath,amsfonts,amssymb}
\usepackage{graphicx}
\usepackage{setspace}
\usepackage{tocloft}
\usepackage{threeparttable}
\usepackage{tabularx}
\usepackage{booktabs}
\title{Transplant-Ready? Evaluating AI Lung Segmentation Models in Candidates with Severe Lung Disease}

\author[a]{Jisoo Lee}
\author[a]{Michael R. Harowicz}
\author[c*]{Yuwen Chen}
\author[c]{Hanxue Gu}
\author[d]{Isaac S. Alderete}
\author[a]{Lin Li}
\author[a,b,c,e]{Maciej A. Mazurowski}
\author[f]{Matthew G. Hartwig}

\affil[a]{Department of Biostatistics and Bioinformatics, Duke University, Durham, NC, 27703, USA}
\affil[b]{Department of Radiology, Duke University, Durham, NC, 27703, USA}
\affil[c]{Department of Electrical and Computer Engineering, Duke University, Durham, NC, 27703, USA}
\affil[d]{School of Medicine, Duke University, Durham, NC, 27703, USA}
\affil[e]{Department of Computer Science, Duke University, Durham, NC, 27703, USA}
\affil[f]{Department of Surgery, Duke University, Durham, NC, 27703, USA}

\cftpagenumbersoff{figure}
\cftpagenumbersoff{table} 
\begin{document} 
\maketitle

\begin{abstract}
\textbf{Purpose}: This study evaluates publicly available deep-learning based lung segmentation models in transplant-eligible patients to determine their performance across disease severity levels, pathology categories, and lung sides, and to identify limitations impacting their use in preoperative planning in lung transplantation.

\textbf{Approach}: This retrospective study included 32 patients who underwent chest CT scans at Duke University Health System between 2017 and 2019 (total of 3,645 2D axial slices). Patients with standard axial CT scans were selected based on the presence of two or more lung pathologies of varying severity. Lung segmentation was performed using three previously developed deep learning models: Unet-R231, TotalSegmentator, MedSAM. Performance was assessed using quantitative metrics (volumetric similarity, Dice similarity coefficient, Hausdorff distance) and a qualitative measure (four-point clinical acceptability scale). 

\textbf{Results}: Unet-R231 consistently outperformed TotalSegmentator and MedSAM in general, for different severity levels, and pathology categories (p$<$0.05). All models showed significant performance declines from mild to moderate-to-severe cases, particularly in volumetric similarity (p$<$0.05), without significant differences among lung sides or pathology types. 

\textbf{Conclusions}: Unet-R231 provided the most accurate automated lung segmentation among evaluated models with TotalSegmentator being a close second, though their performance declined significantly in moderate-to-severe cases, emphasizing the need for specialized model fine-tuning in severe pathology contexts.
\end{abstract}


{\noindent \footnotesize\textbf{*} Yuwen Chen,  \linkable{yuwen.chen@duke.edu} }

\begin{spacing}{2}   

\section{Introduction}
\label{sect:intro}  
Donor-recipient size matching is crucial for successful lung transplantation, to ensure the donor lung fits within a fixed thoracic cavity. The current lung allocation system incorporates recipient height as a surrogate for thoracic cavity size to improve size matching \cite{alderete2025thoracoabdominal}, however, this approach remains imprecise and disproportionately affects shorter candidates. Traditionally, predicted total lung capacity (pTLC), derived from height-based equations \cite{eberlein2013donor},serves as a proxy for alveolar space. Yet, pTLC fails to account for pathologic alterations of the thoracic cavity—such as hyperinflation, fibrosis, or effusions—which frequently occur in patients with end-stage lung disease and often distort the anatomical confines of the thoracic cavity.  Since size matching directly influences both graft selection and postoperative outcomes \cite{barnard2013size}, there is increasing demand for more precise, anatomy-based metrics.

Computed tomography (CT) volumetry has emerged as a reliable and practical solution for preoperative thoracic cavity assessment, offering direct measurements that are unavailable through spirometry or anthropometry \cite{prabhu2024computed}. Furthermore, CT is routinely obtained during the transplant evaluation process, enabling opportunistic use for volumetric analysis without requiring additional imaging or testing \cite{gauthier2019chest}. For single lung transplants, CT allows for targeted assessment of the transplant hemithorax, critical in patients with asymmetric pathology or prior surgical interventions. Additionally, prior work has demonstrated that CT-derived lung volumes correlate more strongly with actual TLC than pTLC, and outperform pTLC in predicting post-transplant survival \cite{prabhu2024computed,jung2016feasibility}. However, accurate CT volumetry hinges on proper segmentation of thoracic structures. Manual segmentation remains the gold standard, but is time-consuming, labor-intensive, operator-dependent, and infeasible at scale \cite{mcgrath2020manual}. Automated segmentation, particularly deep learning-based models, offers a faster and more reproducible alternative. Publicly available models—such as Unet-R231 \cite{hofmanninger2020automatic}, TotalSegmentator \cite{wasserthal2023totalsegmentator}, and MedSAM \cite{ma2024segment}—have demonstrated strong performance in lung CT analysis, but their application in LTx remains limited and largely invalidated. 

Despite these advantages, applying automated segmentation models in LTx recipients presents several challenges. First, deep learning-based models often struggle in the setting of severe pathology—such as dense consolidations, emphysema, fibrosis, and pneumothorax—which are common among LTx candidates. Model performance depends more heavily on the diversity and quality of training datasets than on model architecture \cite{hofmanninger2020automatic}. However, most publicly available algorithms were developed using datasets containing normal or mildly diseased lungs. Second, segmentation boundaries remain inconsistently defined: some models exclude collapsed or non-aerated lung regions, fail to differentiate solid versus ground glass-opacities, or ignore the presence of pleural air and fluid—each of which impacts total volume calculations and leads to inaccurate size matching. These challenges highlight the need for rigorous validation of publicly available automated lung segmentation models to assess their reliability in LTx.

Accordingly, this study evaluates the performance of publicly available automated segmentation models in transplant-eligible candidates considering different lung sides, disease categories, and severity strata. We hypothesize that current models, while promising in more general populations, will demonstrate variable performance in the pathologically distorted lungs of LTx candidates, limiting their immediate clinical applicability. We assess each model’s accuracy and clinical interpretability, as well as key limitations that may impact pre-operative planning. By identifying model limitations, this study offers insights into improving segmentation methods for LTx applications.

\section{Methods \& Materials}
This retrospective study was approved by the Duke University Institutional Review Board (Pro00106785) with a waiver of informed consent. The study followed the Checklist for Artificial Intelligence in Medical Imaging (CLAIM) guidelines. 

\subsection{Dataset}
This study was designed to evaluate the performance of deep learning-based lung segmentation models across a wide range of lung pathologies with varying severity. 

\begin{figure}
    \centering
    \includegraphics[width=0.8\linewidth]{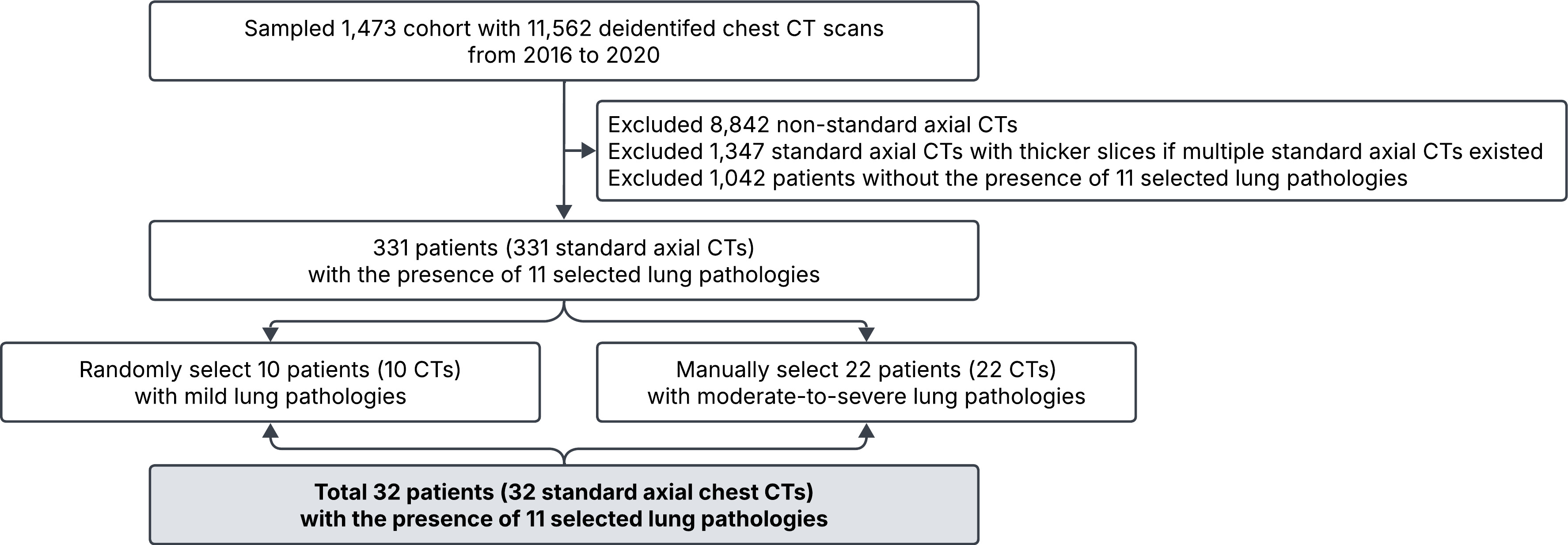}
    \caption{Inclusion/exclusion criteria}
    \label{fig:criteria}
\end{figure}

\begin{figure}
    \centering
    \includegraphics[width=1\linewidth]{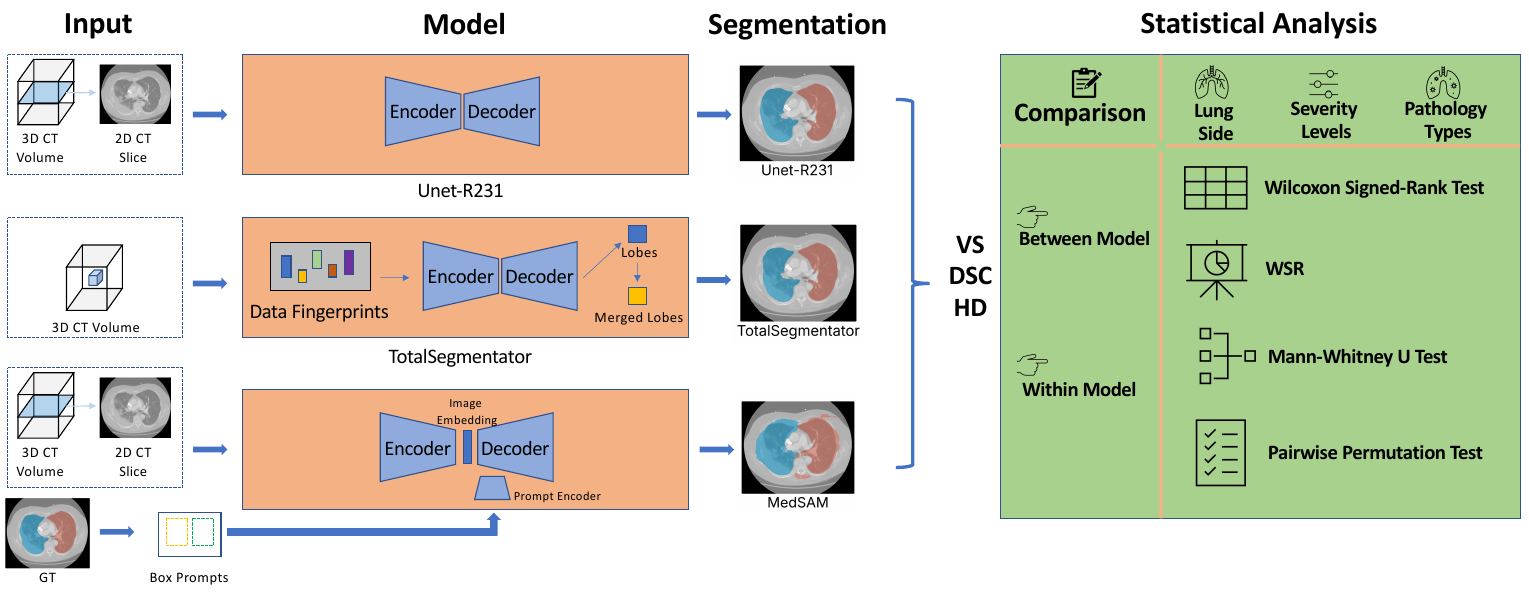}
    \caption{Experiment pipeline}
    \label{fig:pipeline}
\end{figure}

The dataset was derived from patients who underwent chest CT scans at Duke Health System between 2016 and 2020. Exams were deidentified, and we used honest broker service within Duke’s secure Protected Analytics Computing Environment (PACE). Standard axial chest CT scans were chosen for their high anatomical resolution and widespread clinical use in lung assessment. The SeriesDescription field from the DICOM header was used to identify the appropriate imaging sequence. If multiple standard axial sequences were available for a patient, the scan with the larger slice thickness was selected to reduce annotation burden. Patients were included based on the presence of the following lung pathologies in radiology reports: atelectasis, consolidation, ground-glass opacities, interlobular septal thickening, nodules, masses, emphysema, fibrosis (reticular, bronchiectasis, distortion, ground-glass), pleural effusion, pneumothorax, and pleural thickening/nodularity. These lung pathologies were chosen for their clinical significance, prevalence in LTx scenarios, and diagnostic feasibility on CTs. The severity of each lung abnormality was categorized as either mild (encompassing mild, small, or few abnormalities) or moderate-to-severe (moderate-to-severe, medium-to-large, or multiple-to-many) by M.R.H. (a fellowship-trained cardiothoracic radiologist with 2 years of post-fellowship experience). A patient was classified as moderate-to-severe if any of their lung abnormalities were assessed as moderate-to-severe. Two sampling strategies were used: (1) random selection of mild cases (n=10) and (2) manual selection of moderate-to-severe cases (n=22). The final dataset included 32 standard axial chest CT scans (3,645 2D axial slices)  in NifTI format from 32 patients with two or more of the 11 lung pathologies. The inclusion/exclusion criteria and evaluation pipeline are outlined in Fig. \ref{fig:criteria} and \ref{fig:pipeline}.

\subsection{Annotation}
In this study, lungs were defined as the entire pleural space, including lung tissue and any associated fluid or abnormalities, while excluding extra-thoracic structures such as mediastinum, chest wall, and external air pockets. The reference standard was set by the consensus of M.G.H. ($\#\#$ years of experience in LTx) and M.R.H. Initial annotations were manually annotated by J.L. (no formal radiology training) using 3D Slice \cite{fedorov20123d}, and then were reviewed and approved by M.R.H. to ensure accuracy. Left and right lungs were individually labeled.

\begin{table*}[h]
\tiny
\centering
\begin{threeparttable}
\caption{Details of three publicly available lung segmentation models}
\label{tab:model_details}
\renewcommand{\arraystretch}{1.2}
\begin{tabularx}{\textwidth}{l|X|X|X}
\toprule
\textbf{Feature} & \textbf{Unet-R231} & \textbf{TotalSegmentator} & \textbf{MedSAM} \\
\midrule
Backbone model
  & U-Net
  & nnU-Net
  & SAM \\
\midrule
Target task
  & 2D lung CT segmentation
  & 3D whole-body CT segmentation
  & 2D multi-modal medical image segmentation \\
\midrule
Training dataset$^{*}$
  & Chest CTs from clinical routine practice
  & Whole-body CTs from clinical routine practice
  & Publicly available medical image datasets \\
\midrule
\# of patients
  & 231
  & 1{,}082
  & Not provided \\
\midrule
\# of CT scans
  & 231
  & 1{,}082
  & 6{,}226 \\
\midrule
\# of CT scans presenting lungs
  & 231
  & Not provided
  & Maximum 1{,}564$^{**}$ \\
\midrule
\# of CT scans presenting lung pathologies
  & Minimum 55
  & Not provided
  & Not provided \\
\midrule
Lung definition
  & Left and right lungs individually
  & Left and right lungs individually, combining lobes for each side
  & Bounding-box prompts based on lung masks from public datasets \\
\midrule
Exclusion
  & Trachea
  & Trachea$^{***}$
  & Not specified \\
\midrule
Inclusion
  & Air pockets, tumour, and pleural effusions
  & Air pockets, tumour, and pleural effusions$^{***}$
  & Not specified \\
\midrule
Lung pathologies
  & Severe pathologies (e.g., consolidations, effusions, fibrosis, ground-glass opacity, tumors, trauma)
  & No specific lung abnormalities; includes five broad abnormal categories (bleeding, inflammation, tumor, vascular)
  & No specific lung abnormalities \\
\midrule
Annotation method for lungs
  & Manual refinement of auto-segmentations using U-Net trained on VISCERAL Anatomy3 dataset (18)
  & Manual refinement of auto-segmentations using U-Net–Lobe (19)
  & Masks from publicly available datasets \\
\midrule
Codes available
  & ---
  & \href{https://github.com/wasserth/TotalSegmentator}{github.com/wasserth/TotalSegmentator}
  & \href{https://github.com/bowang-lab/MedSAM}{github.com/bowang-lab/MedSAM} \\
\midrule
\bottomrule
\end{tabularx}
\begin{tablenotes}[flushleft]
\footnotesize
\item $^{*}$ Training dataset only, not including validation or test datasets.
\item $^{**}$ Estimated from the combination of training CT datasets where segmentation targets are whole-body or abdominal organs and lungs.
\item $^{***}$ Not specified in the paper; information derived from the source paper of the model used for annotation.
\end{tablenotes}
\end{threeparttable}
\end{table*}

\subsection{Deep Learning-based Lung Segmentation Models}
Three publicly available deep learning-based lung segmentation models were chosen for their popularity and relevance to lung segmentation tasks. Detailed characteristics of these models are presented in Table \ref{tab:model_details}.

Unet-R231 is a classic 2D Unet model trained on 231 chest CTs from clinical routine, involving 55 cases with severe lung pathologies, such as consolidations, effusions, fibrosis, ground-glass opacity, tumors, and trauma. It processed 3D images as a series of 2D slices and segmented left and right lungs separately.

TotalSegmentator (version 2) is a 3D nnUnet model, an advanced Unet–based architecture that automatically configures hyperparameters based on dataset characteristics \cite{wasserthal2023totalsegmentator}. It was trained on 1,082 whole-body CTs, part of which contained lungs. Although its training dataset did not specifically focus on lung abnormalities, it included various abnormalities. TotalSegmentator initially segmented lobes for each side and subsequently merged them to generate left and right lung masks.

TotalSegmentator (version 2) is a 3D nnUnet model, an advanced Unet–based architecture that automatically configures hyperparameters based on dataset characteristics \cite{isensee2021nnu}. It was trained on 1,082 whole-body CTs, part of which contained lungs. Although its training dataset did not specifically focus on lung abnormalities, it included various abnormalities. TotalSegmentator initially segmented lobes for each side and subsequently merged them to generate left and right lung masks.

MedSAM is a fine-tuned Segment-Anything Model (SAM) \cite{kirillov2023segment} based on a 2D vision transformer model utilizing user-defined box prompts to guide segmentation—for multi-modal medical imaging. It was trained on diverse public medical image datasets, including 6,226 CTs with various segmentation targets. MedSAM is substantially different than the two models described above. It does not fully automatically segment lungs in the CTs but rather requires the users to provide a prompt indicating where the lungs are. In this study, MedSAM processed 3D images as a series of 2D slices, with box prompts defined based on ground-truth lung masks.

For each model, its original code was used without modification for image pre-processing and lung segmentation to ensure reproducibility. Experiments were performed on an AMD EPYC 7302 (16-Core) with 512 GB of RAM and 1 NVIDIA A6000 GPU.

\subsection{Evaluation Metrics}
To evaluate automated lung segmentation models, we employed both quantitative and qualitative metrics. 

Quantitative evaluation used three primary metrics—Volumetric Similarity (VS), Dice Similarity Coefficient (DSC), and Hausdorff Distance (HD)—to objectively measure segmentation accuracy. These metrics were calculated between the predicted segmentation masks and the ground-truth masks in 3D for each case. For each metric, we reported the mean, standard deviation, and $95\%$ confidence intervals (CIs) with 1,000 bootstrapping, providing a comprehensive view of model performance. Detailed definitions and formulas for these metrics can be found in Method S1 in Supplement.

Qualitative evaluation (hereinafter referred to as clinical quality rate or CQR) incorporated expert radiologist assessment (M.R.H.) on 3D volume to ensure clinical relevance. Segmentation quality was rated based on clinical acceptability using a four-point scale:
\begin{itemize}
    \item 1 (Poor): Errors account for a change in a single lung volume (left or right) of $> 15\%$, or if a pathology is completely or nearly completely missed by the segmentation.
    \item 2 (Acceptable): Errors account for a change in a single lung volume (left or right) of $5-15\%$.
    \item 3 (Good): Errors account for a change in a single lung volume (left or right) of $< 5\%$.
    \item 4 (Excellent): Fully acceptable with no clinical errors.
\end{itemize}

This qualitative assessment complements quantitative metrics by addressing clinically significant issues that may not be reflected in numerical scores, such as missing pathological features or including non-lung structures.

\subsection{Statistical Analysis}
Overall performance differences across models and within models were evaluated based on lung side, severity level, and pathology category, using statistical tests tailored to each comparison type. The Shapiro-Wilk \cite{shapiro1965analysis} and the Levene’s \cite{gastwirth2009impact} tests were used to check normality and variance equality. For overall between-model comparisons on the same lung side or severity level, the Wilcoxon signed-rank test \cite{woolson2007wilcoxon}, a non-parametric test for paired data, was used, with Holm-Bonferroni correction \cite{holm1979simple} applied to maintain strong Type I error control across multiple comparisons. Overall, within-model comparisons used the Wilcoxon signed-rank test for paired lung side comparisons and the Mann-Whitney U test \cite{nachar2008mann} for independent severity level comparisons. For overall between-model comparisons on the same pathology type, pairwise permutation test (10,000 permutations) with Holm-Bonferroni correction was applied due to the small sample size in some types. Overall within-comparisons among pathology types were assessed using independent permutation tests (10,000 permutations) with Benjamini-Yekutieli correction \cite{benjamini2001control} to account for partial patient overlap across types. Pathology subgroup analysis was also performed: between-model comparisons used pairwise permutation tests with Holm-Bonferroni correction, while within-model comparisons relied on independent permutation tests. A two-tailed P-value of $<0.05$ was considered statistically significant. Statistical analyses were performed using Python (version 3.11) with SciPy (version 1.15) and statmodel (version 0.14) libraries.

\section{Results}
\begin{table}[h]
\tiny
\centering
\resizebox{0.7\linewidth}{!}{%
\begin{threeparttable}
\caption{Characteristics of the test dataset}
\label{tab:test_dataset}
\renewcommand{\arraystretch}{1.2}
\begin{tabular}{ll}
\toprule
\textbf{Variable} & \textbf{Dataset (no. of patients)} \\
\midrule
Age (y, mean $\pm$ SDs) & 60.31 $\pm$ 15.58 \\
\midrule
\textbf{Sex} & \\
Female & 10 \\
Male & 22 \\
\midrule
\textbf{Race} & \\
White/Caucasian & 23 \\
Black/African American & 7 \\
Multiracial & 1 \\
Not reported & 1 \\
\midrule
\textbf{Severity Levels} & \\
Mild & 10 \\
Moderate-to-Severe & 22 \\
\midrule
\textbf{Pathologies (total, (mild: moderate-to-severe))} & \\
\quad Atelectasis & 24 (21:3) \\
\quad Consolidation & 13 (8:5) \\
\quad Emphysema & 12 (5:7) \\
\quad Fibrosis & 10 (6:4) \\
\quad Ground-glass opacities & 19 (13:6) \\
\quad Interlobular septal thickening & 7 (0:7) \\
\quad Mass & 3 (3:0) \\
\quad Nodule & 23 (20:3) \\
\quad Pleural effusion & 16 (11:5) \\
\quad Pleural thickening/nodularity & 5 (0:5) \\
\quad Pneumothorax & 4 (1:3) \\
\bottomrule
\end{tabular}
\end{threeparttable}
}
\end{table}

This study included 32 patients (22 males, 10 females; mean age $60.3 \pm 15.6$ years) who underwent chest CT scans for various lung pathologies. The dataset encompassed a broad range of lung pathologies, with the most prevalent being atelectasis (n=24), nodules (n=23), and ground-glass opacities (n=19). Less frequent pathologies included pneumothorax (n=4) and masses (n=3). Most patients (n=22) had moderate-to-severe disease, while a smaller group (n=10) exhibited only mild abnormalities. This diversity in pathology types and severity levels ensured a clinically representative cohort for segmentation model evaluation. Detailed data characteristics were presented in Table \ref{tab:test_dataset}.

\begin{figure}
    \centering
    \includegraphics[width=1\linewidth]{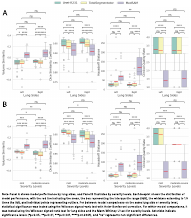}
    \caption{Overall comparison of segmentation performance across models by lung side and severity level.}
    \label{fig:comparison}
\end{figure}

\subsection{Overall Comparison Split by Lung Side}
Based on Fig. \ref{fig:comparison}A, Unet-R231 consistently demonstrated the best performance, achieving the closest-to-zero VS, highest DSC, lowest HD, and highest CQR. TotalSegmentator followed closely, with performance differences compared to Unet-R231 reaching significance for both lungs in VS and DSC (p$<$0.05 and p$<$0.0001 each), but not in HD or CQR. MedSAM lagged across all metrics with significantly lower accuracy and higher variability (mostly p$<$0.0001). Within models, performance was slightly reduced in the left lung compared to the right, though this was only statistically significant for MedSAM in HD (p$<$0.05).

\subsection{Overall Comparison Based on Severity Level}
As shown in Fig. \ref{fig:comparison}B, Unet-R231 remained the top performer and significantly outperformed TotalSegmentator in VS and DSC across both severity levels (p$<$0.001); however, no significant differences were observed in HD or CQR between the two. All models showed decreased performance in moderate-to-severe cases compared to mild. This within-model declines were significant for Unet-R231 in VS (p$<$0.05) and CQR (p$<$0.001), and for TotalSegmentator in VS (p$<$0.01), DSC (p$<$0.05), and CQR (p$<$0.01). MedSAM’s performance was consistently poor and showed no significant differences between severity levels.

\begin{figure}
    \centering
    \includegraphics[width=1\linewidth]{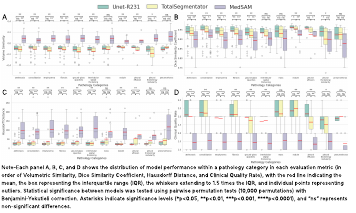}
    \caption{Overall between-model comparison of segmentation performance by pathology category.}
    \label{fig:comparisoncate}
\end{figure}

\subsection{Overall Pathology Category Comparison}
\begin{figure}[h]
    \centering
    \includegraphics[width=1\linewidth]{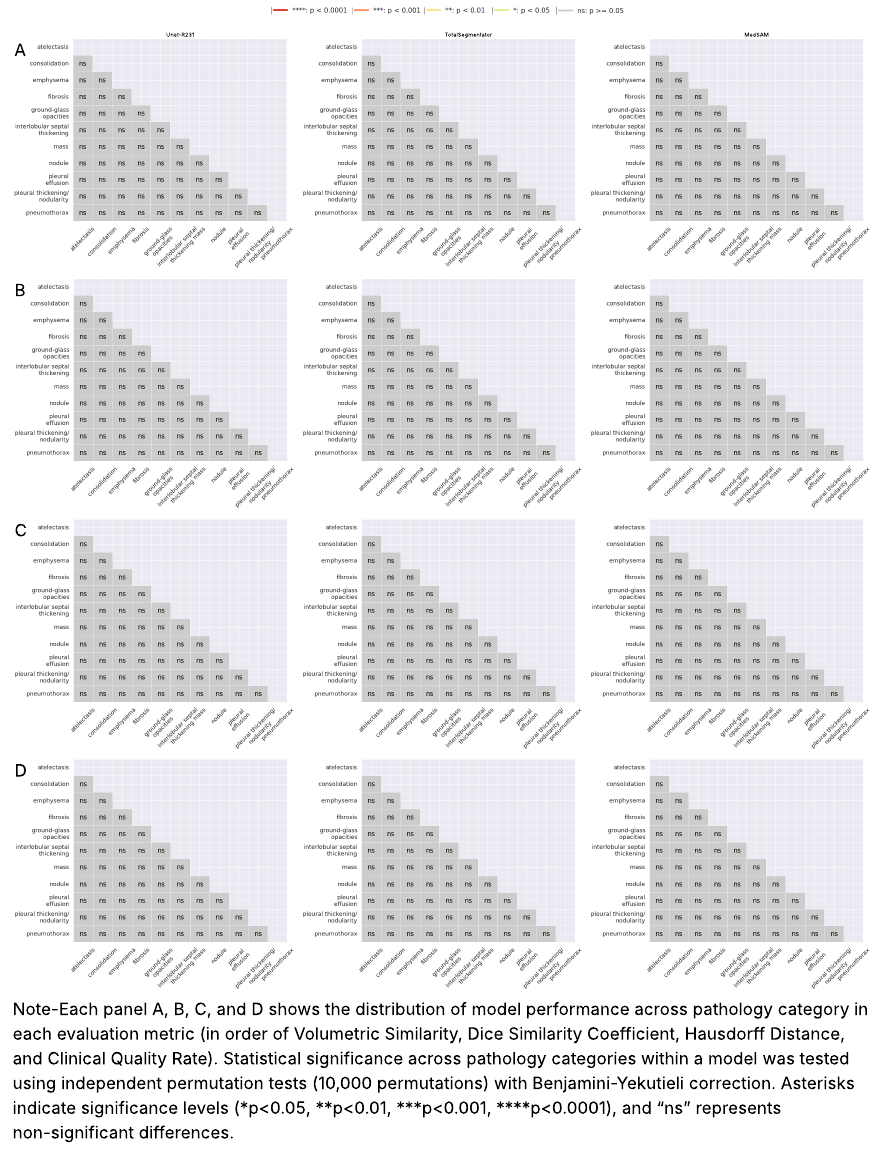}
    \caption{Overall within-model comparison of segmentation performance by pathology category.}
    \label{fig:s1}
\end{figure}

According to Fig. \ref{fig:comparisoncate}, Unet-R231 and TotalSegmentator maintained stable performance across most pathologies. Significant differences between these models emerged only in a few categories, such as fibrosis (VS, p$<$0.01) and four categories in DSC (e.g., fibrosis p$<$0.01; ground-glass opacities, nodule, and pleural thickening/nodularity p$<$0.05). HD and CQR showed no significant differences between the two. MedSAM consistently performed worse across nearly all pathologies and metrics (p$<$0.01), except in mass and pneumothorax, where no between-model differences were detected. Within each model (Fig. \ref{fig:s1}), segmentation performance did not significantly vary across pathology types (p$>$0.05).

\subsection{Subgroup Lung Side Comparison}
\begin{figure}[h]
    \centering
    \includegraphics[width=1\linewidth]{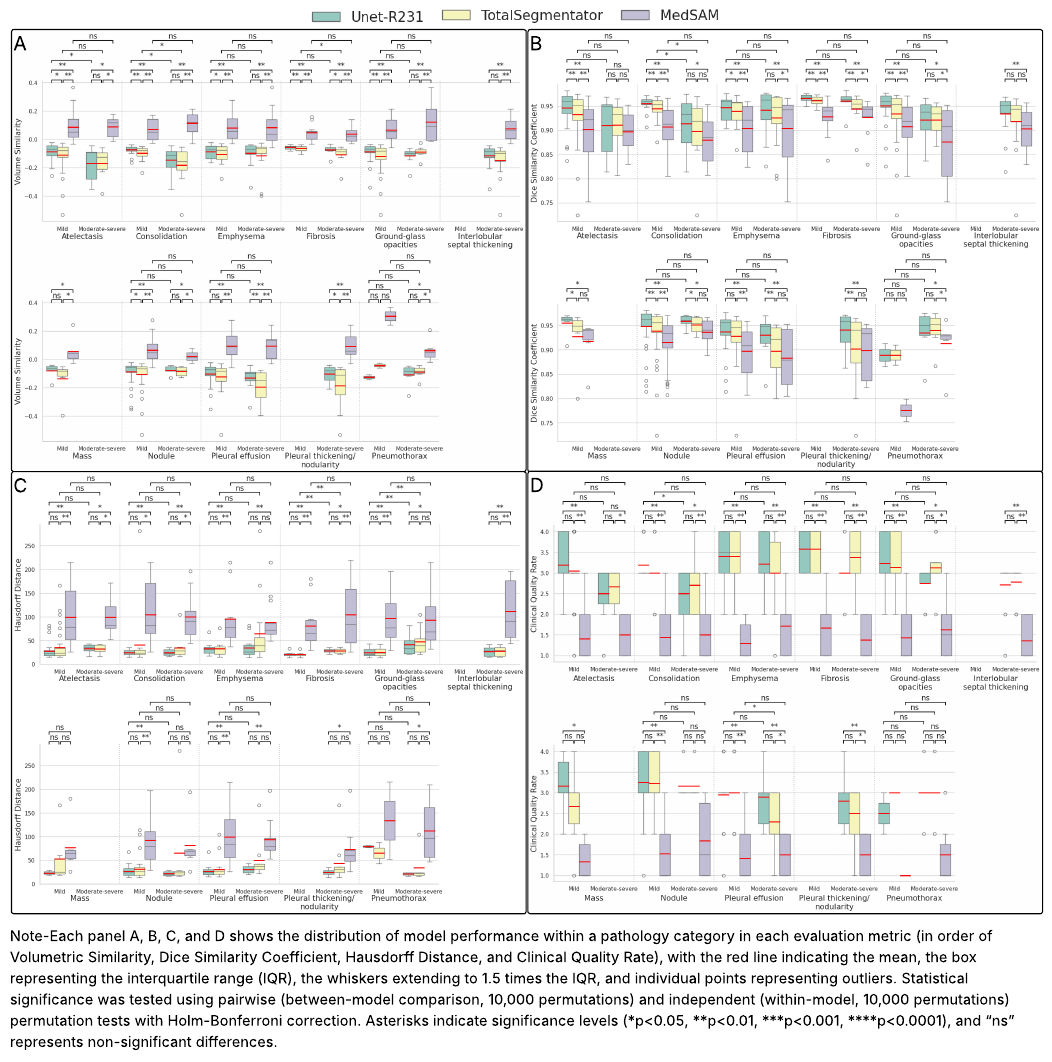}
    \caption{Pathology subgroup comparison of segmentation performance across models by lung sides.}
    \label{fig:s2}
\end{figure}

Based on Fig. \ref{fig:s2}, under most pathology categories, Unet-R231, TotalSegmentator and MedSAM exhibited consistent segmentation performance between left and right lungs, with no significant side-to-side differences in VS, DSC, HD and CQR. Only MedSAM showed a significant underperformance on the right lung in HD (p$<$0.01) for pneumothorax, despite its VS, DSC and CQR remaining comparable across both sides.

\begin{figure}
    \centering
    \includegraphics[width=1\linewidth]{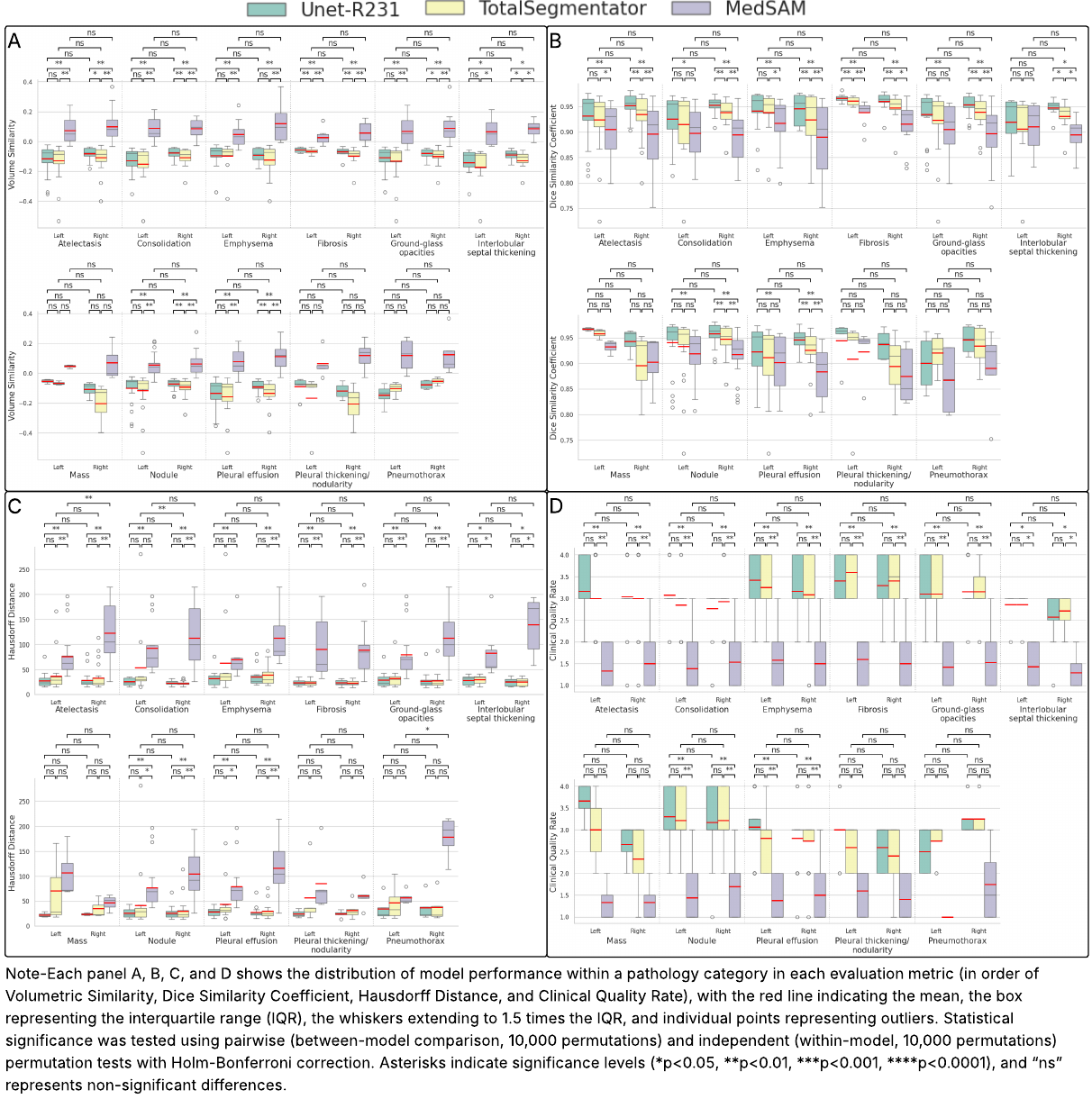}
    \caption{Pathology subgroup comparison of segmentation performance across models by severity levels.}
    \label{fig:subgroup}
\end{figure}

\subsection{Subgroup Severity Level Comparison}
As shown in Fig. \ref{fig:subgroup}, models have exhibited superior performance at a mild level compared with moderate-to-severe level. At the mild severity level, Unet-R231 achieved significantly higher VS for atelectasis and consolidation, while TotalSegmentator outperformed for consolidation and fibrosis (all p$<$0.05). For the DSC, both Unet-R231 and TotalSegmentator showed superior performance in consolidation (p$<$0.05). In terms of HD, each model performed significantly better in fibrosis and ground-glass opacities (all p$<$0.01). Finally, for the CQR, Unet-R231 significantly outperformed in consolidation (p$<$0.05).

\section{Discussion}
This study evaluated three publicly available deep learning-based models for automated lung segmentation in transplant-eligible patients. Our findings confirm that models developed for general populations do not consistently perform well in the pathologically distorted lungs commonly seen in LTx candidates. Although Unet-R231 and TotalSegmentator showed relatively strong overall performance, both experienced notable performance declines in moderate-to-severe disease—particularly in VS and CQR, two metrics critical for preoperative planning. These results underscore a key limitation: even high-performing, publicly available models are vulnerable to segmentation errors in anatomically complex lungs.

Consistent with prior research, these findings highlight the importance of training dataset composition in determining model robustness. Unet-R231’s superior performance likely stems from its training on chest CT datasets enriched with severe lung abnormalities, making it more adaptable to complex imaging appearances typical of LTx candidates. TotalSegmentator was trained on a broader set of whole-body CTs, with fewer severely diseased lungs, likely contributing to its slightly lower accuracy. In contrast, MedSAM’s poor performance across all metrics reflects a fundamental limitation in both its training approach and model design, as it was developed using general medical datasets, which may dilute its capability on CTs, and relies on user-defined prompts rather than pathology-aware segmentation.

Importantly, this study also suggests that disease severity is an influential determinant of model reliability. While none of the models exhibited significant performance differences across specific pathology types, both Unet-R231 and TotalSegmentator showed consistent declines in segmentation accuracy with increasing disease severity. Given that most LTx candidates present with moderate-to-severe disease, these results emphasize the need for models that are optimized specifically for this subgroup to ensure reliable clinical application.

Furthermore, our results suggest that deep learning-based models can provide consistent segmentation across left and right lungs, supporting their use in side-specific transplant planning. However, both Unet-R231 and TotalSegmentator consistently produced slightly smaller segmentation volumes than ground truth (negative VS), likely due to the exclusion of peripheral and collapsed lung regions. This systematic under-segmentation could lead to underestimation of thoracic cavity volume and ultimately compromise donor-recipient size matching—an essential factor in transplant success.
This study has several limitations. First, the relatively small sample size (32 patients) and the imbalance in pathology type distribution may limit the generalizability of our findings. We addressed this by applying non-parametric tests and permutation methods to ensure robust statistical analyses. Second, all CTs were acquired at a single institution, but variations in imaging protocols and scanner types were not controlled, potentially impacting model performance. Third, potential confounders, such as prior surgical interventions or underlying comorbidities, were not explicitly accounted for, though the diverse pathology inclusion mitigates some bias.

Despite these limitations, our findings offer significant clinical implications, particularly for lung transplantation and similar clinical contexts involving severe lung pathologies. First, future models should be trained or fine-tuned on datasets rich in moderate-to-severe pathologies to ensure robustness in complex cases. Second, standardized segmentation definitions—including clear guidelines on whether to include peripheral, non-aerated, or fluid-filled regions—are necessary to avoid underestimation in volumetric analyses.

\section{Conclusion}
This study evaluated three publicly available deep learning-based lung segmentation models—Unet-R231, TotalSegmentator, and MedSAM—on 32 transplant-eligible patients across varying lung pathologies, severity levels, and lung sides. Quantitative and qualitative assessments showed Unet-R231 performed best overall, but all models experienced significant accuracy declines in moderate-to-severe disease and exhibited systematic under-segmentation. These findings underscore the limitations of current models in complex pathological contexts and highlight the need for fine-tuning on severity-enriched datasets and establishing standardized segmentation protocols for reliable preoperative lung transplantation planning.

\bibliography{report}   
\bibliographystyle{spiejour}   

\clearpage

\section{Appendix: Supplementary Material}
Method S1. Detailed definitions and formulas for quantitative evaluation metrics
Volumetric Similarity ($VS$) quantifies the volume agreement between predicted ($V_A$) and ground truth ($V_B$) volumes: $VS=2(V_A-V_B )/(V_A+V_B)$. Negative VS indicates the predicted volume is underestimated, while positive VS denotes the predicted volume is overestimated. This metric is particularly valuable in preoperative lung assessment, where precise volume estimation is essential for transplant suitability. 

Dice Similarity Coefficient ($DSC$) measures the spatial overlap between predicted ($A$)  and ground truth ($B$) segmentations: $DSC = \frac{2|A \cap B|}{|A| + |B|}$. A DSC of 1 indicates perfect overlap, while 0 denotes no overlap. This metric is widely used in medical image segmentation tasks.

Hausdorff Distance ($HD$) evaluates the boundary accuracy by calculating the maximum Euclidean distance ($D(a,b)$) between boundary pixels of predicted ($A$) and ground truth ($B$) segmentations: $H(A,B) = \max_{a \in A} \left\{ \min_{b \in B} D(a,b) \right\}$. A lower HD indicates better boundary alignment, which is essential for cases involving lung abnormalities that require precise delineation for surgical planning.

\clearpage

\begin{figure}[h]
    \centering
    \includegraphics[width=1\linewidth]{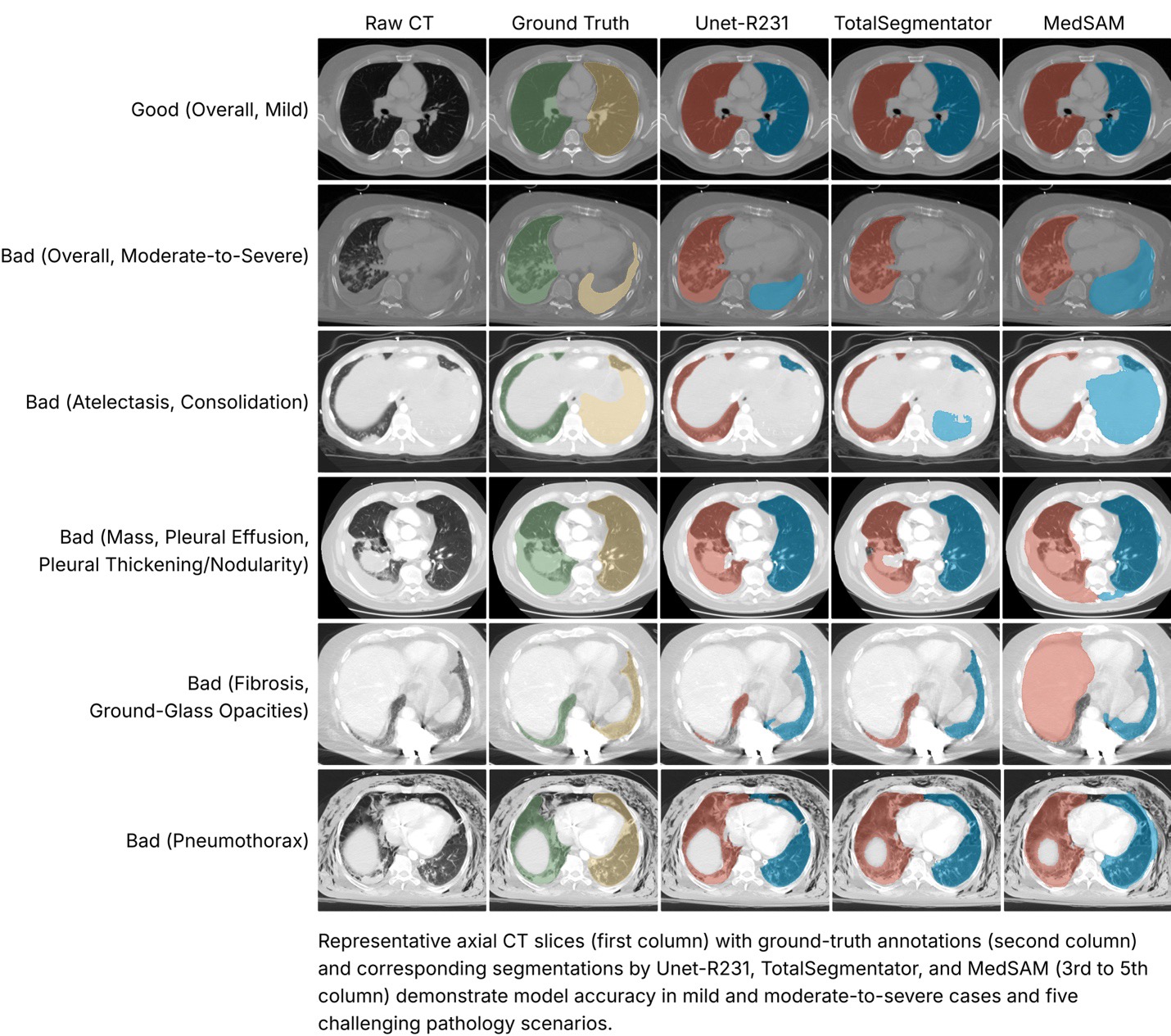}
    \caption{Examples of lung segmentation model performance between severity levels and pathology categories.}
    \label{fig:s3}
\end{figure}

\clearpage

\subsection*{Disclosures}
The authors declare that there are no financial interests, commercial affiliations, or other potential conflicts of interest that could have influenced the objectivity of this research or the writing of this paper.

\subsection* {Code, Data, and Materials Availability} 
Data and code generated or analyzed during the study may be available from the corresponding author by request.







\end{spacing}
\end{document}